\documentclass{article}
\usepackage{spconf,amsmath,graphicx}
\usepackage{amsmath,latexsym,graphicx,verbatim}
\usepackage{amssymb}
\usepackage{subfigure}
\usepackage{multirow,color}

\usepackage{caption}
\usepackage{hyperref}
\usepackage[noadjust]{cite}

\title{Non-contact photoplethysmogram and instantaneous heart rate estimation from infrared face video \thanks{Work partially supported by DoD, NIH, NSF, Google, Cisco, and Microsoft.}}

\name{Natalia Martinez$^{\star}$ \qquad Martin Bertran$^{\star}$ \qquad Guillermo Sapiro$^{\star}$ \qquad Hau-Tieng Wu$^{\dagger}$}

\address{$^{\star}$ Department of Electrical and Computer Engineering, Duke University \\
    $^{\dagger}$ Department of Mathematics and Department of Statistical Science, Duke University}

\begin{document}
%
\maketitle
\begin{abstract}
 Extracting the instantaneous heart rate (iHR) from face videos has been well studied in recent years. It is well known that changes in skin color due to blood flow can be captured using conventional cameras. One of the main limitations of methods that rely on this principle is the need of an illumination source. Moreover, they have to be able to operate under different light conditions. One way to avoid these constraints is using infrared cameras, allowing the monitoring of iHR under low light conditions. In this work, we present a simple, principled signal extraction method that recovers the iHR from infrared face videos. We tested the procedure on 7 participants, for whom we recorded an electrocardiogram simultaneously with their infrared face video. We checked that the recovered signal matched the ground truth iHR, showing that infrared is a promising alternative to conventional video imaging for heart rate monitoring, especially in low light conditions. Code is available at \url{https://github.com/natalialmg/IR_iHR}.
\end{abstract}
%

\section{Introduction}
\label{sec:intro}

The gold standard for monitoring instantaneous heart rate (iHR) is electrocardiogram (ECG) \cite{dawson2010changes}. Another popular noninvasive technique is photoplethysmogram (PPG) \cite{alian2014photoplethysmography,allen2007photoplethysmography}. Both techniques require direct skin contact with the subject, which might not be suitable in contexts such as driver drowsiness, or sleep monitoring. PPG relies on measuring the rapid variations in light absorption in an illuminated skin region caused by the difference in absorption curves for oxigenated and non-oxigenated blood. This principle motivated the use of digital cameras to measure the plethysmographic signals from face videos under ambient light conditions \cite{verkruysse2008remote,poh2010non, davila2016physiocam}. Several methodologies for estimating heart rate from face videos have been developed over the years \cite{poh2011advancements,kwon2012validation,li2014remote,kumar2015distanceppg,lam2015robust,tulyakov2016self}. In particular, \cite{wang2018comparative} provides a comprehensive overview of the history of the research done in this area and compares the performance of some of these approaches. As a general rule, most of these methods need an illumination source, depend on color band
manipulation, and require control over the signal acquisition process (e.g., controlled light sources, or subjects remaining motionless during acquisition).

The recent inclusion of infrared (IR) cameras in many conventional devices, coupled with their resilience to low-light and variable-light conditions, make them especially attractive for remote monitoring in the context of iHR detection. Their use has just now started to be explored in the detection of heart rate using infrared face videos \cite{chen2016realsense,zhang2018heart}, but so far these approaches are limited to estimating a heart rate average over a considerable time frame (over 30 seconds). This paper shows that, under controlled motion conditions, it is feasible to extract even sub-second approximations to the iHR using basic spatiotemporal analysis and time-frequency analysis.

We describe this approach, and show its performance on face IR videos acquired using a Kinect camera from 7 healthy volunteers. The extracted iHR is compared against ECG and contact PPG ground truth signals that were simultaneously acquired.



\section{Non-contact PPG signal from IR video}
\label{sec:ihrfrir}

Here we describe the proposed algorithm to construct the non-contact PPG signal from an IR face video and hence extract the instantaneous heart rate. We divide this process into three main steps. The first step is detecting and segmenting the face in the video into $n_r$ disjoint spatial regions. 
Secondly, we take the mean activity of each region, denoise it, and decompose it into a smaller subset of sources. Finally, we introduce a signal quality index to select the signals of interest, and combine them to construct the non-contact PPG signal we are after. Figure \ref{fig:pip} summarizes the initial preprocessing stages, while Figure \ref{fig:svd} shows an example of the subsequent recovery process of the non-contact PPG.

\subsection{Input IR video}

For each subject, denote the recorded IR video as $V:\mathbb{R}\to \mathbb{R}^{n\times m}$, where $V(t)$ denotes the recorded frame at time $t$, which is of size $n\times m$ (height $\times$ width). Suppose the video is sampled every $\tau$ seconds; that is, sampled at $1/\tau$ Hz, and the recording starts at time $0$ and lasts for $T$ seconds. We have thus $n_t=\lfloor T/\tau\rfloor$ frames. In this study, $1/\tau = 58 Hz$. We additionally assume that the subject's head is fixed, so major movements between frames are ignored.

\subsection{Preprocessing the IR video}

\begin{figure}[h!]
\centering
 \begin{tabular}{cc}
 \centering
 \subfigure[Detect facial landmarks
			on IR video.]{%
 \includegraphics[height=0.25\textwidth]{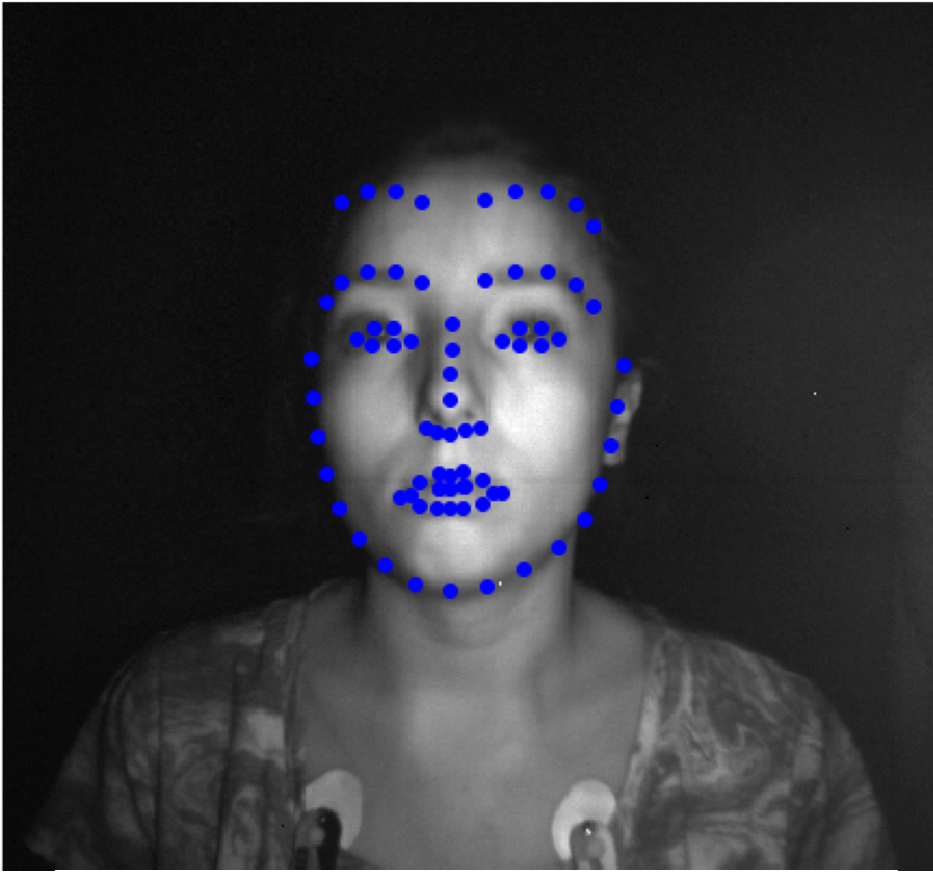}}
 \subfigure[Segment face into regions.]{%
 \includegraphics[height=0.25\textwidth]{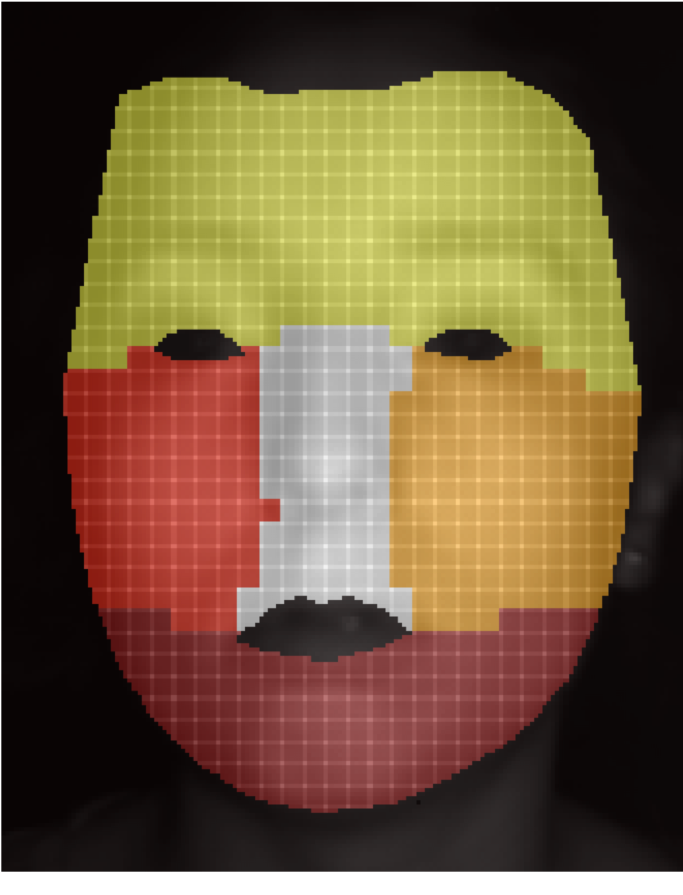}}
 \end{tabular}

	\caption{Outline of video preprocessing pipeline. Figure a) shows how the facial landmarks are detected using Dlib. Figure b) shows how the detected face is subdivided using a 5x5 pixel grid; these regions can be grouped into 5 major facial areas. The mean activity signal of all grid elements compose our observation matrix $Y_0$.}
	\label{fig:pip}
\end{figure}

We detect the boundaries of the face using the Dlib landmark detector \cite{dlib09} on the average face location, frame-by-frame detection is not performed since the subject is assumed to be immobile. An example is shown on Figure \ref{fig:pip}(a). We then divide the area inside the detected face into $n_r$ disjoint regions following a predefined mesh grid. 

Denote those disjoint regions as $R_i$, $i=1,\ldots,n_r$. For each video frame $V_j:=V(j\tau)\in \mathbb{R}^{n\times m}$, $j=1,\ldots,n_t$, we compute the mean IR value on each region $R_i$
\[
y_{i,j} := \frac{1}{|R_i|}\sum_{(x,y)\in R_i} V_j(x,y).
\]
As a result, we obtain the data matrix 
\[
Y_0 \in \mathbb{R}^{n_r \times n_t}.
\] 
In other words, the $i$-th row of matrix $Y_0$ contains a time series with the mean IR activity over region $R_i$, there are $n_r$ such regions defined across the face. We will refer to these as channels. Note that the constructed data matrix is commonly encountered in spatiotemporal analysis. For the purposes of this study, the face is subdivided into regions using a non-overlapping $5\times5$-pixel grid. See Figure \ref{fig:pip}(b) for illustration.

To denoise the time series, we apply to each channel in the data matrix $Y_0$ an order $5$ bandpass Butterworth with cutoff frequencies at $24$ and $300$ bpm, a range that comfortably acommodates most normal heart rates. Denote the filtered signals as the data matrix $Y$. This bandpass filter is chosen based on the physiological knowledge that, for a normal subject, the heart rate is between $40$ and $200$ bpm.

\subsection{Low rank spatiotemporal model}

We assume that the IR video captures different physiological dynamics, such as respiration, body movement, and hemodynamics, among others. Denote these physiological sources as $X_i \in \mathbb R^{n_t}$, where $i=1,\ldots,n_s$ and $n_s \leq \min\{n_r,n_t\}$. Note that in general $X_i$ and $X_j$ might not be orthogonal when $i\neq j$; for example, the hemodynamics and respiration might be coupled due to the respiratory sinus arrhythmia.

The data matrix $Y$ is then modeled as a mixture of these $n_s$ source signals with additive and uncorrelated noise
\begin{equation}
Y = AX + \sigma Z,
\end{equation}
where $X \in \mathbb R^{n_s \times n_t}$ contains the physiological source signals, $A \in \mathbb R^{n_r \times n_s}$ is the source mixture matrix, $Z$ is a noise matrix with independent and identically distributed entries with zero mean, unit variance and finite fourth moment, and $\sigma^2>0$ is a scalar constant that describes the noise variance. In other words, the recorded signal on each region, $Y_i$, is a mixture of different sources via $A$, contaminated by noise. We further make the low rank assumption that $n_s$ is fixed and small. This assumption means that there are limited sources of physiological dynamics that are captured by the IR video. 

\subsection{Determine important sources}

Due to the low-rank assumption and the high-dimensional nature of the spatiotemporal model, apply SVD to the data matrix $Y$: 
\begin{equation}
{Y} = U\Lambda V, 
\end{equation}
where $U\in O(n_r)$ consists of the left singular vectors, $V\in O(n_t)$ consists of the right singular vectors, and $\Lambda\in \mathbb{R}^{n_r\times n_t}$ consists of singular values $\sigma_1\geq\sigma_2\geq\ldots\sigma_{\min\{n_r,n_t\}}\geq0$. Denote $u_i$ and $v_i$ to be the $i$-th left and right singular vectors respectively. Note that $V$ contains the relevant temporal signals that are mixed in each region, and $U$ their weight in each spatial location. An example is illustrated in Figure \ref{fig:svd}.

Denote $\beta := n_t/n_s$
and denote $n_s^*$ is the number of singular values such that $\eta(\sigma_i/\sigma)>0$.
Since the noise level $\sigma$ is in general not known, we estimate it as proposed in \cite{donoho2013optimal}:
\[
\hat{\sigma} = \frac{\textup{median}(\vec{\sigma_i})}{\sqrt{\mu_b}},
\]
where $\mu_b$ is the median of the Marcenko-Pastur distribution \cite{marchenko1967distribution} with parameter $\lambda_{\pm} = (1\pm\sqrt{\beta})^2$. Applying this procedure to $Y$ reduced the number of non-zero singular values by over $80\%$ on average.

\subsection{Reconstructing the non-contact PPG signal}

Due to the non-orthogonal nature of physiological sources, we cannot recover $X$ directly from $Y$ by applying the usual blind source separation technique. We thus propose the following procedure to reconstruct the non-contact PPG signal.

Define a signal quality index (SQI) for a signal $x$ of length $n_t$ as
\[
Q(x) := \frac{\int_{\frac{3}{4}f_{p} }^{\frac{5}{4}f_{p} } |\hat{x}(f)| df}{\int_{\frac{1}{2}f_{p} }^{2f_{p} } |\hat{x}(f)| df},
\]
where $\hat{x}$ is the Fourier transform of the time series $x$, and $f_p$ is the expected heart rate of a normal subject. Note that $Q(x)$ quantifies how concentrated the time series $x$ is around $f_p$ in the frequency domain. 

\begin{figure}[h!]
	\centering
 \includegraphics[clip,width=0.45\textwidth]{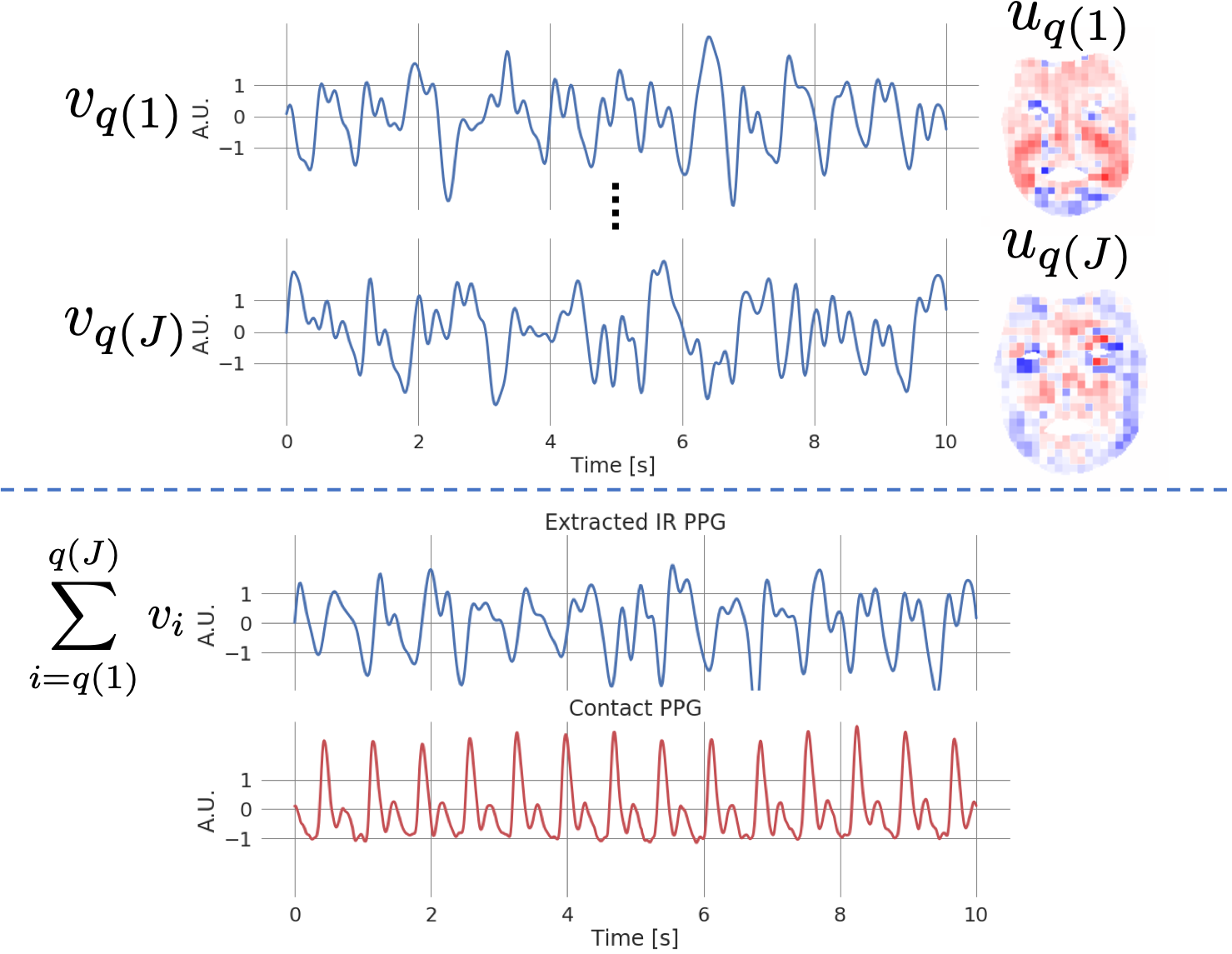}

	\caption{Top half of figure shows left ($u$) and right ($v$) singular vectors sorted by SQI in descending order. The resulting accumulated non-contact PPG ($\textup{PPG}_{\texttt{IR}}$) is shown on the bottom, ground truth contact PPG is shown for comparison. Contact and noncontact PPG show well matched cycles.}
	\label{fig:svd}
\end{figure}

\begin{figure}[h!]
	\centering
 \includegraphics[clip,width=0.45\textwidth]{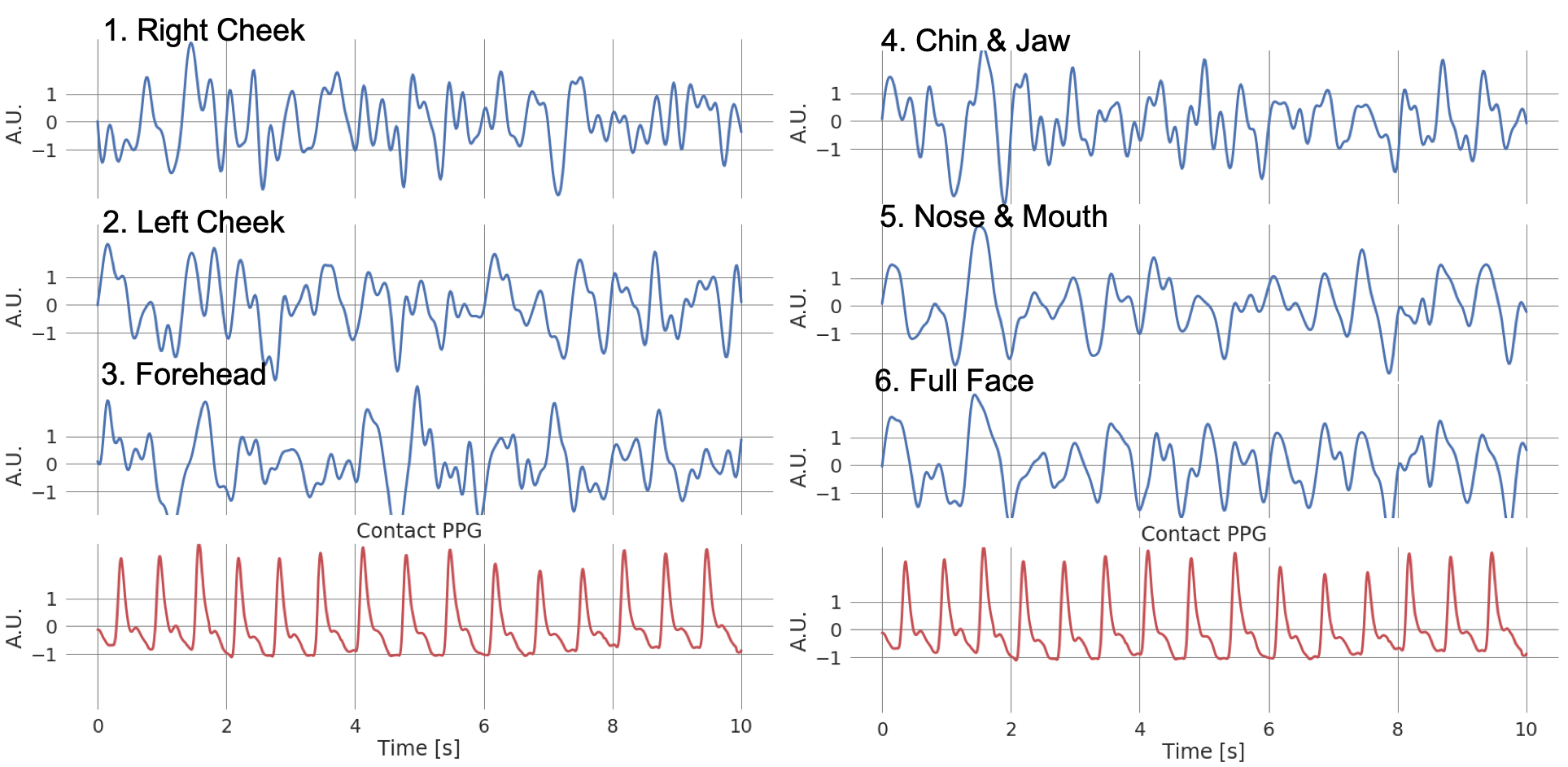}

	\caption{Results of applying the proposed framework to the channels contained in each of the 5 major facial areas independently. All areas contain a measure of iHR information, the best results are obtained by analyzing the full face as a whole. Ground truth PPG is provided for comparison.}
	\label{fig:faceRegions}
\end{figure}

\begin{figure}[ht]
	\centering
	\includegraphics[width=0.5\textwidth]{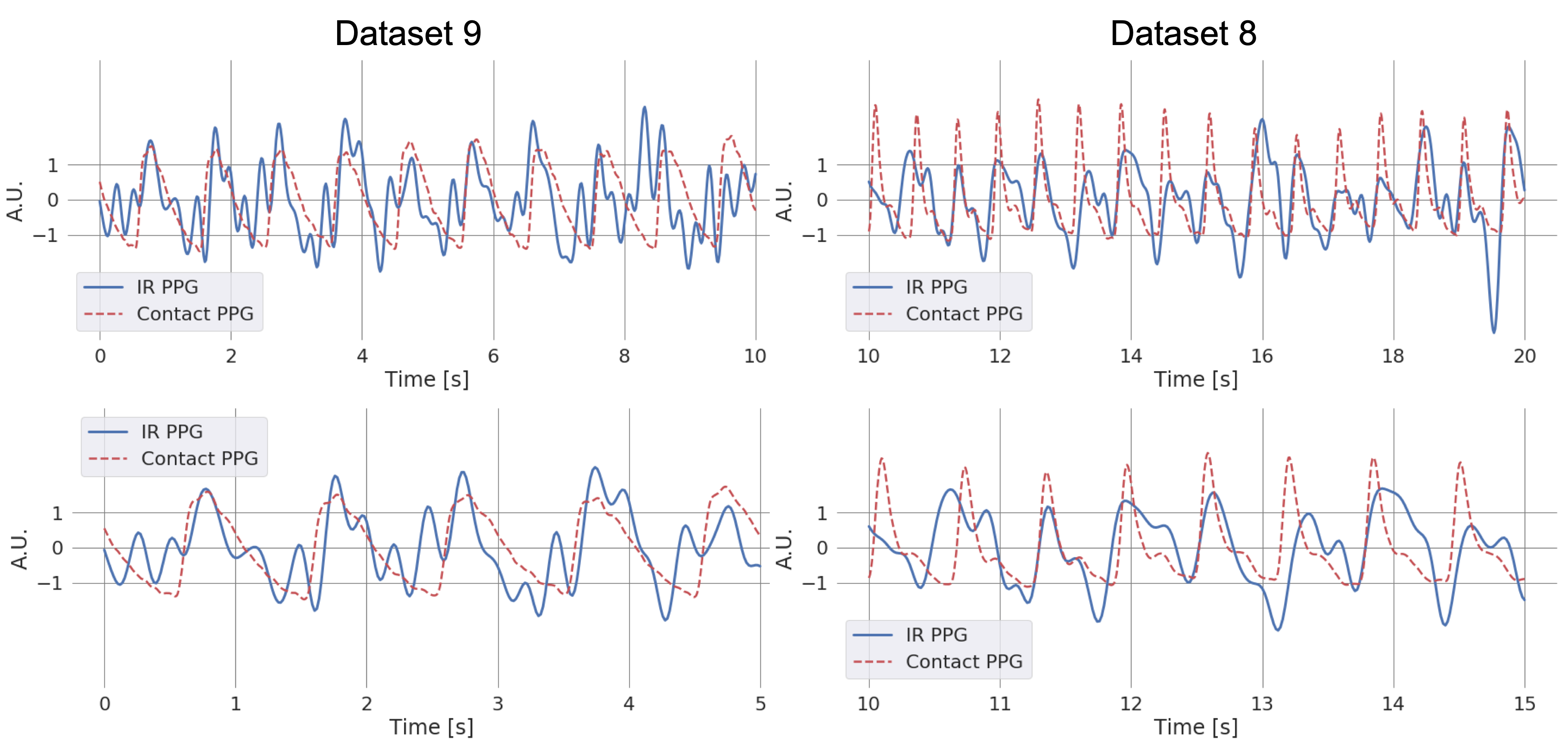}
	\caption{Estimated $\textup{PPG}_{\texttt{IR}}$ (solid blue) compared against ground truth contact PPG (dashed red) at $10s$ and $5s$ timescales for two datasets. Overall, cycles are well matched between contact and noncontact PPG. Datasets are taken from subjects with dissimilar resting heartrate.}
	\label{fig:closeup}
\end{figure}

We rank all temporal signals $v_i$, where $i=1,\ldots,n_s^*$, according to their SQIs $Q(v_i)$. Consider the reordering permutation $q:\{1,\ldots,n_s^*\}\to \{1,\ldots,n_s^*\}$ so that $Q_{q(1)} \geq Q_{q(2)}\geq \ldots$. Our hemodynamic estimator, the non-contact PPG signal denoted as $\textup{PPG}_{\texttt{IR}}$, is defined as \vspace{-.1cm}
\[
\textup{PPG}_{\texttt{IR}} := \sum_{i = q(1)}^{q(J)} v_i\in \mathbb{R}^{n_t},
\]
for $J\in \mathbb{N}$ chosen by the user.
Here we determine $J$ by greedily accumulating the sources until the maximal quality is achieved; that is,
\vspace{-.1cm}
\[ 
q(J) = \arg\max_{j} Q\left(\sum_{i = q(1)}^{q(j)}v_i\right).
\]
Figure \ref{fig:svd} shows an outline of the recovery process for non-contact PPG over the full face. Figure \ref{fig:faceRegions} shows the recovered iHR signal when we applied the proposed method to the channels contained in each of the five major facial areas independently, this is provided merely for illustration purposes. In general, using the entire facial area provided the best results. Figure \ref{fig:closeup} shows short time segments of non-contact PPG compared against ground truth contact PPG. Additional examples will be provided in the following sections.
\vspace{-.1cm}
\subsection{Estimation of the instantaneous heart rate}

Denote the short time Fourier transform (STFT) of the constructed non-contact PPG signal as $S_{\textup{PPG}_{\texttt{IR}}}\in \mathbb{C}^{n_t\times (n_t/2)}$, where $S_{\textup{PPG}_{\texttt{IR}}}(t,f)$ is the STFT coefficient at time $t/\tau$ and frequency $f/T$.
%
%
From the STFT we extract the dominant curve using the curve extractor proposed in \cite{cicone2017nonlinear},
\begin{align}
\hat{c}=& \arg\max_{c\in \mathbb{N}^{n_t}} \sum_{t=1}^{n_t} \log|S_{\textup{PPG}_{\texttt{IR}}}(t,c(t))|\\
 &\quad-\lambda \sum_{t=2}^{n_t}|c(t)-c(t-1)|)\in \mathbb{N}^{n_t} ,\nonumber
\end{align}
\vspace{-.1cm}
where $\lambda>0$ is a regularization constant. The iHR is thus determined by 
\[
\textup{iHR}:=\hat{c}/T\in \mathbb{R}^{n_t}.
\]
Figure \ref{fig:results} shows the obtained $\textup{PPG}_{\texttt{IR}}$ and its STFT; ground truth iHR from ECG is also shown for comparison.

\section{Experiments}

We acquired 9 simultaneous ECG, PPG, and IR face video using a standard patient monitor (Philips IntelliVue MP70 Patient Monitor) and a Microsoft Kinect camera. The clocks in the Kinect camera and the patient monitor were synchronised with a time accuracy of $\pm1s$. Acquisitions were done over 7 healthy subjects. The subjects were asked to look straight into the camera and maintain a steady posture, but otherwise behave, blink, and breathe normally. The instantaneous heart rate (iHR) was estimated from the IR video using the process described in Section \ref{sec:ihrfrir}. Ground truth iHR was extracted from the ECG signal using the R-peak detection algorithm implemented in the python library biosppy \cite{biosppy}.

\section{Results}

\begin{figure}[ht]
	\centering
	\includegraphics[width=0.48\textwidth]{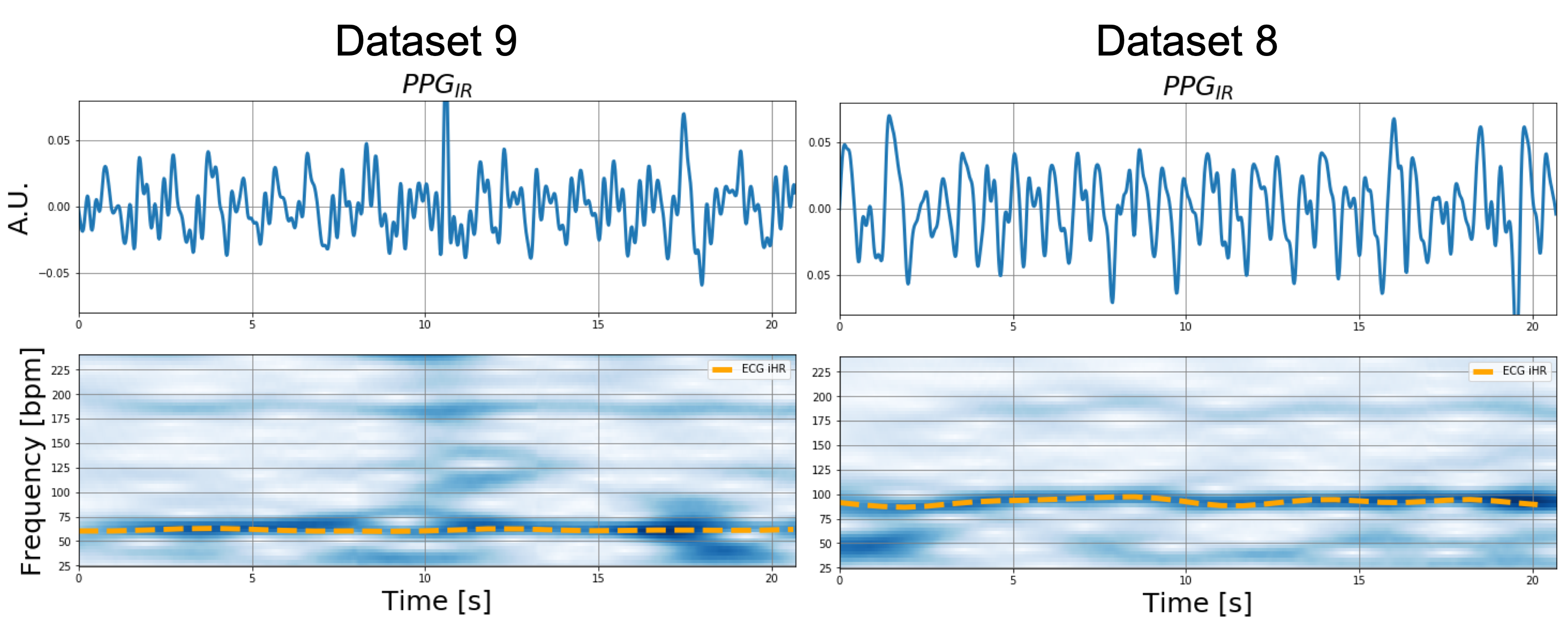}
	\caption{Recovered noncontact PPG ($\textup{PPG}_{\texttt{IR}}$) and its corresponding STFT are shown for dataset 9 and 8. Dotted orange line shows the ground truth iHR recovered from ECG signal. Spectrograms show good correspondence between recovered IR iHR and ground truth iHR.}
	\label{fig:results}
\end{figure}

For each of the 9 datasets we measured the differences between the recovered iHR signal and ground truth using root mean square error (RMSE) and relative error \[\frac{1}{n_t}\sum^{n_t}_{t=1} \frac{|iHR(t) -iHR_{ECG}(t)|}{iHR_{ECG}(t)}.\]
Table \ref{tab:table1} shows these values. Figure \ref{fig:results} shows the extracted iHR signals. Implemented code is available at \url{https://github.com/natalialmg/IR_iHR}.


			
	

\begin{table}[h!]
\begin{center}
\begin{tabular}{lllll}
\hline
\multicolumn{1}{c}{\multirow{2}{*}{\textbf{\begin{tabular}[c]{@{}c@{}}Dataset /\\ Subject\end{tabular}}}} & \multicolumn{3}{c}{\textbf{RMSE {[}bpm{]}}} & \multicolumn{1}{c}{\textbf{\begin{tabular}[c]{@{}c@{}}Relative\\ error {[}\%{]}\end{tabular}}} \\ \cline{2-5} 
\multicolumn{1}{c}{} & Every 1s & Every 10s & Every 30s & Every 30 s \\ \hline
d1/1 & 5.39 & 4.27 & 4.03 & 4.50 \\
d2/1 & 5.61 & 4.99 & 5.22 & 6.51 \\
d3/2 & 4.71 & 4.44 & 3.70 & 4.40 \\
d4/2 & 3.59 & 2.87 & 1.33 & 1.56 \\
d5/3 & 4.39 & 3.86 & 1.95 & 2.60 \\
d6/4 & 4.95 & 4.65 & 2.91 & 3.58 \\
d7/5 & 2.21 & 1.31 & 1.02 & 1.60 \\
d8/6 & 3.30 & 1.42 & 0.23 & 0.25 \\
d9/7 & 2.38 & 1.26 & 0.66 & 1.08 \\ \hline
\end{tabular}
\end{center}
\caption{Error measures across datasets}
\label{tab:table1}
\end{table}

In general, RMSE results averaged for longer time-frames ($30s$) are satisfactory. Perhaps surprisingly, RMSE results for iHR at $1s$ intervals are also reasonable. Figure \ref{fig:results} shows good correspondence between the ground truth ECG iHR and the STFT of the recovered non-contact PPG.

\section{Concluding remarks}

In this paper, we extracted non-contact PPG from IR facial video. We showed that a simple, principled method based on matrix decomposition was sufficient to recover instantaneous heart rate with small relative errors on a second-by-second basis when subjects remain relatively stationary. 

This suggests the viability of IR for non-contact PPG, particularly when we consider the low-light and varying-light performance of IR in general compared to traditional RGB methods. Additional work is required to adequately and robustly correct for motion artifacts. Improvements can be done on the process by which we combine the singular vectors to obtain our final hemodynamic estimator. Finally, more research needs to be done on the characterization of absorption curves of biological processes of interest in the near infrared spectrum.
We leave this physiological research as a future collaborative work. We could also consider more sophisticated time-frequency representation tools to further analyze the obtained non-contact PPG signal for the instantaneous heart rate estimation. A more general manifold learning algorithm and matrix denoise technique can be applied to capture motion and time latency; for example, due to the high dimensional nature of $Y$,
the matrix $Y$ can be denoised by the optimal shrinkage algorithm proposed in \cite{gavish2017optimal}: 
$\tilde{Y} = \sum_{i=1}^{n_s^*} \sigma\eta(\sigma_i/\sigma )u_i v_i^T$,
where $\eta(y)= \left\{
\begin{array}{ll} 
\frac{\sqrt{(y^2-\beta -1)^2 -4\beta}}{y} & y > 1+\sqrt{\beta}\\
0 & y \le 1+\sqrt{\beta}
\end{array}\right.$
is the optimal shrinkage under the Frobenius norm. This approach has the potential to further improve the overall quality of the signal.
We will explore these possibilities in future work. 

\bibliographystyle{IEEEbib}
\bibliography{refs}

\end{document}